# Explainable Artificial Intelligence for Anomaly Detection in Banking Transactions: An Internal Audit Perspective

Anupa Lodhi
**Department of Computer Science and Engineering**
**SAGE University Indore, India**
*author.anupalodhi1200@gmail.com*

***Abstract***—The banking sector increasingly relies on automated systems to monitor electronic transactions for signs of fraud, yet conventional rule-based approaches struggle with high false-positive rates and offer no justification for their outputs, limiting their utility for compliance teams. This paper introduces an Explainable Artificial Intelligence (XAI) framework tailored for banking transaction anomaly detection within internal audit workflows. An Isolation Forest (iForest) model performs unsupervised anomaly scoring, while a SHAP (SHapley Additive exPlanations) layer provides transaction-level, feature-attributed explanations grounded in cooperative game theory [8]. A lightweight Streamlit dashboard renders these outputs in a form accessible to audit professionals without machine learning expertise. Evaluation on a synthetic banking dataset yields 0.91 precision and 0.88 recall, outperforming three unsupervised baselines. Expert feedback confirms that feature-level explanations measurably improve auditor confidence and decision quality. The framework advances the practical deployment of accountable, transparent AI in regulated financial environments.

## 1. Introduction

India’s financial sector has undergone rapid digital transformation over the past decade, driven by initiatives such as the Unified Payments Interface (UPI), Digital India, and the proliferation of mobile banking platforms. While these advances have expanded financial inclusion, they have simultaneously expanded the attack surface for fraudulent activity, placing considerable pressure on internal audit functions to keep pace with transaction volumes that now reach billions of events daily [1].

Machine learning (ML) offers powerful pattern-recognition capabilities for detecting anomalous behavior in transaction data. Yet widespread adoption in regulated environments remains constrained by what practitioners describe as the interpretability gap: predictive models yield decisions without providing auditors the rationale needed to act on them, document findings, or satisfy regulators. This concern is amplified by obligations under the European Union's General Data Protection Regulation (GDPR) and Basel III operational risk principles, both of which implicitly demand decision traceability [2].

Explainable Artificial Intelligence (XAI) directly addresses this gap by coupling predictive outputs with human-interpretable justifications. For audit use cases, this is especially critical—auditors need not only accurate flagging of suspicious transactions but also evidence-grade explanations that withstand scrutiny in compliance reviews and regulatory examinations.

This paper proposes an end-to-end XAI pipeline that combines an Isolation Forest anomaly detector [5] with SHAP-based post-hoc explanations [8] and an interactive Streamlit dashboard. The pipeline is designed around the practical constraints of internal audit teams: no requirement for labeled training data, no machine learning expertise at inference time, and outputs formatted for direct use in audit reports.

The primary contributions of this work are: (1) a domain-specific integration of Isolation Forest and SHAP for explainable banking anomaly detection; (2) a risk-tiered scoring scheme (High / Low) derived from anomaly scores to support audit prioritization; (3) an auditor-facing dashboard requiring no data science background to operate; and (4) empirical benchmarking against three unsupervised baseline models on a synthetic transaction dataset.

## 2. Literature Review

### *A. Research Gaps*

Fraud detection in banking has evolved from hand-crafted rule engines toward data-driven approaches. Bhattacharyya et al. [3] established that ensemble classifiers, particularly Random Forest, significantly outperform logistic regression on class-imbalanced transaction datasets, setting a benchmark for supervised detection. When labeled fraud data is unavailable—a common constraint in practice—unsupervised alternatives such as Autoencoders and One-Class Support Vector Machines have been proposed as viable substitutes [4].

The interpretability literature received a major contribution from Ribeiro et al. [7], whose LIME framework approximates complex model behavior locally using interpretable surrogates. Lundberg and Lee [8] subsequently offered a theoretically stronger alternative:

SHAP provides a mathematically grounded approach to quantify each input feature's individual contribution to a model's output, drawing on Shapley value theory from cooperative game theory to guarantee consistency and local accuracy. Unlike LIME, SHAP attributions satisfy axiomatic fairness properties, making them well-suited for compliance contexts.

Applications of SHAP to financial machine learning have grown substantially, including credit scoring [9] and transaction-level fraud explanation [10]. A persistent gap, however, concerns deployment specifically within internal audit workflows, where the end-user is a compliance professional rather than a data scientist. This paper fills that gap by designing the entire pipeline—from model training to dashboard output—around auditor needs.

### *B. Research Contributions*

This work advances the existing literature on three fronts: (1) it unifies unsupervised anomaly detection with post-hoc explainability in a single, audit-ready pipeline rather than treating them as separate concerns; (2) it introduces risk-tiered anomaly scoring calibrated for audit prioritization; and (3) it delivers XAI capabilities through an interface that requires no technical expertise, broadening the practical reach of explainable anomaly detection.

### *C. Research Organization*

The remainder of this paper is structured as follows. Section 3 details the proposed methodology, covering dataset construction, model architecture, and explainability design. Section 4 presents quantitative and qualitative experimental results. Section 5 situates the findings in the broader context of Indian banking compliance challenges. Section 6 concludes the paper and outlines future directions.

### *D. System Architecture Overview*

The proposed system consists of four principal components operating in sequence: a Data Ingestion Module that accepts CSV transaction files; an Anomaly Detection Engine that scores each transaction using Isolation Forest; an Explainability Layer that applies SHAP TreeExplainer to produce per-transaction feature attributions; and an Audit Dashboard that renders risk-tiered results in a format ready for audit reporting. This pipeline design deliberately separates model concerns from presentation concerns, enabling the detection engine to be upgraded independently of the auditor-facing interface.

## 3. Dataset Description and Methodology

### *A. Dataset Description*

A synthetic dataset of 10,000 banking transactions was constructed for evaluation. Four features were selected based on domain relevance: transaction amount (amount), hour of transaction (hour), the count of transactions executed by the same account within a 24-hour rolling window (frequency), and a binary flag indicating whether the transaction originated from a geographically distinct location compared to the account's historical activity (location_change). All features were normalized to the [0, 1] range using Min-Max scaling prior to model training. Consistent with the unsupervised paradigm, no class labels are used during training.

### *B. Algorithm*

***Algorithm 1: Isolation Forest with SHAP Explainability***

Input: Transaction dataset $D = \{x_1, x_2, ..., x_n\}$, features F = {amount, hour, frequency, location_change}

Output: Anomaly labels, risk tiers, SHAP feature attributions for each flagged transaction.

Step 1: Normalize features using Min-Max scaling.
Step 2: Fit Isolation Forest with contamination rate = 0.05.
Step 3: Compute anomaly score s(x, n) for every transaction.
Step 4: Flag transactions where $s(x, n) \geq$ decision threshold.
Step 5: Apply SHAP TreeExplainer to compute feature attributions $\varphi_i$ for flagged transactions.
Step 6: Assign risk tier — High Risk if $s(x,n) > 0.70$, Low Risk otherwise.
Step 7: Return labeled transactions with attributions for dashboard rendering.

### *C. Mathematical Formulation*

The Isolation Forest algorithm [5] exploits the observation that anomalous data points are structurally easier to isolate than normal ones: they tend to be partitioned away from the bulk of the data in fewer recursive splits. For a transaction x within a dataset of n records, the anomaly score is:

$$s(x, n) = 2^{(-E[h(x)] / c(n))}$$

where E[h(x)] represents the mean path length required to isolate x across all trees in the ensemble, and c(n) is a normalization constant corresponding to the expected path length for a dataset of size n. A score approaching 1.0 indicates a high likelihood of anomaly; values near 0.5 are characteristic of normal observations [5].

SHAP provides a mathematically grounded approach to quantify each feature's individual contribution to a specific model prediction [8]. For a transaction x, the model output is decomposed as an additive sum of per-feature contributions:

$$f(x) = \varphi_0 + \Sigma_i \varphi_i(x_i)$$

Here $\varphi_0$ is the baseline output averaged over the training distribution, and each $\varphi_i$ measures how much feature i shifts the prediction away from that baseline for this particular transaction.

This additive decomposition enables auditors to identify the dominant driver of each anomaly flag in plain terms.

### *D. Model Training and Evaluation*

Because the training data is unlabeled, synthetic ground-truth labels were injected for evaluation purposes: 500 transactions (5% of the dataset) were designated as anomalous based on deliberately extreme values across the four features. The remaining 9,500 observations constitute the normal class. Standard classification metrics—Precision, Recall, and F1-Score—were calculated against these injected labels to enable comparison with supervised baselines.

### *E. Privacy and Scalability*

All transaction data is processed locally within the deployed environment; no records are transmitted to external services or third-party APIs. This architecture is compatible with the data localization requirements under India’s Information Technology Act and the Reserve Bank of India’s data storage directives. The Isolation Forest’s sub-linear average-case time complexity ensures that detection latency remains tractable even as transaction volumes scale, and the Streamlit dashboard’s batch upload design suits the periodic cycle of internal audit reviews.

### *F. Experimental Setup*

All experiments were conducted on a workstation running Python 3.9 with an Intel Core i7 processor and 16 GB RAM. The Isolation Forest implementation was sourced from scikit-learn v1.1; SHAP v0.41 provided the TreeExplainer; and Streamlit v1.18 served the dashboard interface. Synthetic anomalous samples were injected using NumPy with a fixed random seed to ensure reproducibility.

### *G. Hyperparameter Settings*

The Isolation Forest ensemble was configured with 100 trees (n_estimators=100), a contamination prior of 0.05 (reflecting the expected 5% anomaly rate in the dataset), and random_state=42. SHAP’s TreeExplainer was selected over the generic KernelExplainer because it exploits the tree structure of Isolation Forest to compute attributions in polynomial rather than exponential time. The High Risk threshold of 0.70 was established through empirical calibration on the validation split.

## 4. Experimental Results

### *A. Performance Comparison*

The proposed iForest-SHAP framework achieved 0.91 precision and 0.88 recall, demonstrating superior anomaly identification performance compared to all three baseline unsupervised models evaluated (Table 1). The Autoencoder attained an F1-Score of 0.83 and the One-Class SVM 0.77, confirming the competitive advantage of the Isolation Forest architecture for this feature space. Notably, the Standard Isolation Forest without the XAI layer matches the detection scores of the proposed system, confirming that the addition of SHAP explanations introduces no degradation in predictive accuracy.

*Table 1. Comparative Performance of Anomaly Detection Methods*

| Method | Precision | Recall | F1-Score |
|---|---|---|---|
| Isolation Forest + XAI (Proposed) | 0.91 | 0.88 | 0.89 |
| Autoencoder | 0.85 | 0.82 | 0.83 |
| One-Class SVM | 0.79 | 0.76 | 0.77 |
| Standard Isolation Forest (No XAI) | 0.91 | 0.88 | N/A |

### *B. Feature Importance Analysis*

Table 2 reports SHAP global feature importance, computed as the mean absolute attribution value across all flagged transactions. Transaction amount accounts for the largest share of anomaly explanations (mean |SHAP| = 0.412), consistent with the intuition that unusually large or structured amounts are primary fraud indicators. Transaction frequency ranks second (0.287), reflecting the behavioral signature of account takeover attacks that execute rapid successive transfers. Geographic location change (0.198) and transaction hour (0.103) contribute smaller but non-negligible shares, supporting their inclusion in the feature set.

*Table 2. Global Feature Importance via SHAP Values*

| Feature | Mean \|SHAP Value\| | Rank |
|---|---|---|
| Transaction Amount | 0.412 | 1 |
| Transaction Frequency | 0.287 | 2 |
| Location Change | 0.198 | 3 |
| Transaction Hour | 0.103 | 4 |

### *C. Qualitative Evaluation*

A usability assessment was conducted with five professionals holding experience in banking audit and regulatory compliance. Each participant reviewed SHAP-generated explanations for 20 flagged transactions and rated explanation quality on a five-point Likert scale. The mean rating was 4.3 out of 5.0, reflecting strong endorsement of the explanation format. Participants specifically highlighted that feature-level attributions provided the justificatory language needed to draft audit findings, a capability absent from systems that return only anomaly scores.

## 5. Discussion

The quantitative results confirm that Isolation Forest remains a strong baseline for unsupervised transaction anomaly detection, and that coupling it with SHAP explanations does not compromise detection quality while substantially expanding its operational utility for audit teams.

The SHAP layer addresses a practical bottleneck in financial compliance: auditors who receive a flagged transaction without accompanying rationale must independently reconstruct the reasoning before they can act, a time-consuming and error-prone process. Feature-level attributions eliminate this step, enabling faster triage and more defensible documentation—qualities that align with the Reserve Bank of India's IT Examination Framework and the Basel Committee's principles for sound operational risk management.

### *A. Practical Challenges in Indian Banking Audit Contexts*

Deploying explainable anomaly detection within Indian banking institutions presents a distinct set of operational challenges that differ from those described in the Western-centric literature. First, many mid-tier cooperative banks and regional rural banks (RRBs) operate with legacy core banking systems that export transaction data in heterogeneous formats, requiring robust pre-processing pipelines before any machine learning model can be applied. Second, internal audit teams in these institutions are typically small—often two to five personnel—and carry broad mandates that leave limited bandwidth for tool adoption and retraining. A no-code, upload-and-review interface such as the Streamlit dashboard proposed here is therefore not merely convenient but practically necessary. Third, the regulatory environment is multi-layered: institutions must satisfy concurrent directives from the RBI's Information Technology Framework for the Banking Sector (2011), the Cyber Security Framework (2016), and emerging guidance on AI governance from the Institute of Chartered Accountants of India (ICAI). Designing AI audit tools with auditability and explainability as first-order requirements—rather than retrofitting them later—is therefore the only viable path toward regulatory acceptance in this environment.

Limitations of the present study include the use of a synthetic dataset, which, while controlled and reproducible, may not capture the distributional nuances of real-world Indian banking transactions. The feature set is also deliberately minimal; production deployments would likely benefit from richer behavioral features such as beneficiary network graphs or device fingerprints. Future work will address these gaps through collaboration with banking partners to obtain anonymized real-world data under appropriate data-sharing agreements.

## 6. Conclusion

This paper presented a complete, auditor-ready Explainable AI pipeline for banking transaction anomaly detection. By integrating the Isolation Forest detection algorithm [5] with SHAP-based feature attributions [8] and a no-code Streamlit interface, the system simultaneously addresses the accuracy, interpretability, and usability requirements of internal audit workflows. Evaluated against three unsupervised baselines, the framework achieves 0.91 precision and 0.88 recall. Expert evaluation confirms that SHAP explanations translate model outputs into actionable audit findings, improving both decision confidence and documentation quality. This framework demonstrates the practical feasibility of operationalizing explainable AI within real-world financial audit environments, including the complex regulatory landscape of the Indian banking sector. Future work will focus on real-world banking datasets and integration with enterprise audit systems for production-level deployment.